\documentclass[10pt,a4paper,logo]{googledeepmind}

\usepackage{natbib}
\usepackage{subcaption}

\newcommand{\kvshotblue}{\textcolor[RGB]{79,174,211}}

\usepackage{scalerel} % for logo
\usepackage{pifont}       % 特殊符号
\graphicspath{{figures/}}
\usepackage[capitalize]{cleveref}

\title{
\scalerel*{\includegraphics{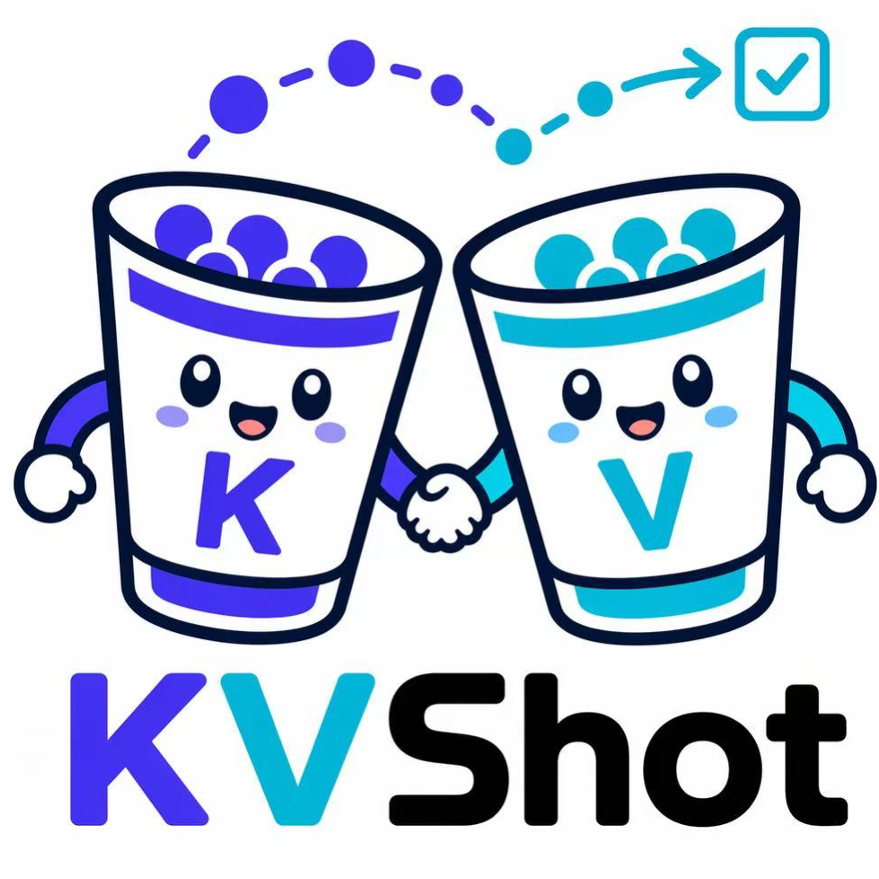}}{{\rule{2.2ex}{2.2ex}}}
When Hidden States Drift: Can KV Caches Rescue Long-Range Speculative Decoding?}
\author[1,2]{Tianyu Liu}
\author[1,3]{Yuhao Shen}
\author[1$\dagger$]{Xinyi Hu}
\author[1]{Baolin Zhang}
\author[1]{Hengxin Zhang}
\author[1]{Jun Dai}
\author[1,3*]{Jun Zhang}
\author[1*]{Shuang Ge}
\author[1]{Lei Chen}
\author[1]{Yue Li}
\author[1]{Mingcheng Wan}
\affil[1]{Qwen Applications Business Group of Alibaba}
\affil[2]{University of Science and Technology of China}
\affil[3]{Zhejiang University}
\makeatletter
\renewcommand\AB@affilsep{\protect\\\protect\Affilfont}
\makeatother
\affil[*]{\begin{minipage}[t]{0.75\textwidth}
Corresponding authors. \quad\mbox{$\dagger$ Project Leader.}\par
\texttt{tianyu\_liu@mail.ustc.edu.cn}, \texttt{\{hxy481617, zj534580, geshuang.zj\}@alibaba-inc.com}
\end{minipage}}

\begin{abstract}
Speculative decoding accelerates large language model inference, but state-of-the-art hidden-state-based drafters (e.g.,~EAGLE3 and MTP) suffer from \emph{\kvshotblue{long-range decay}}: draft accuracy degrades progressively as the speculative step increases. Existing work attributes this decay to train-inference mismatch and proposes autoregressive test-time training (TTT) as a remedy, yet we observe that long-range decay persists even in TTT-trained drafters.
We revisit long-range decay from the perspective of context information preservation. In hidden-state reuse, we argue the target hidden state acts as a \emph{biased context compression}: it aggregates historical token information according to the attention query at the current decoding position, yielding a compact representation optimized for immediate next-token prediction. This compression can suppress information that is less relevant to the current query but becomes important for later speculative steps. In contrast, the target model's KV cache serves as an \emph{explicit context}, retaining the complete set of token-wise KV representations rather than collapsing the history into a single hidden representation. We therefore posit the \textit{KV-Reuse Hypothesis}: allowing the draft model to reuse the target KV cache can provide richer conditioning signals for long-horizon drafting. To test this hypothesis, we introduce \textit{KVShot}, a diagnostic framework that compares three reuse paradigms: \textbf{hidden-only}, \textbf{KV-only}, and \textbf{hybrid}. Extensive evaluations on Qwen3-8B show that KV-Reuse improves long-range acceptance, although end-to-end speedups remain marginal under current training pipelines. Our analysis identifies two key structural bottlenecks: shallow drafters struggle to estimate target queries accurately, and draft-side KV projections receive sparse gradient signals. These findings suggest that realizing the full potential of KV-aware decoding requires moving beyond autoregressive TTT toward block-wise training paradigms. By exposing these bottlenecks, \textit{KVShot} provides a foundational diagnostic testbed and a clear roadmap for designing next-generation inference architectures.

\end{abstract}

\begin{document}
\maketitle

\section{Introduction}
\label{sec:intro}

Autoregressive decoding in large language models (LLMs) is inherently
sequential, making inference latency a persistent bottleneck even with
highly optimized kernels.  Speculative decoding alleviates this by letting
a lightweight draft model propose multiple candidate tokens that a larger
target model verifies in a single forward pass, amortizing the cost of
sequential generation over several tokens at once
\citep{sps1,sps2,specinfer,sequoia,xia2024survey}.

\begin{figure}[t]
    \centering
    \makebox[\textwidth][l]{%
        \hspace*{0.05\textwidth}%
        \begin{subfigure}[t]{0.3\textwidth}
            \centering
            \includegraphics[width=\linewidth]{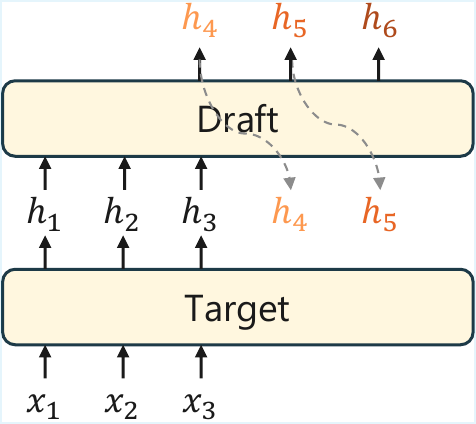}
            \captionsetup{justification=centering,singlelinecheck=true}
            \caption{Hidden-state recursion.}
            \label{fig:long-range-decay-mechanism}
        \end{subfigure}%
        \hspace*{0.14\textwidth}%
        \begin{subfigure}[t]{0.48\textwidth}
            \centering
            \includegraphics[width=\linewidth]{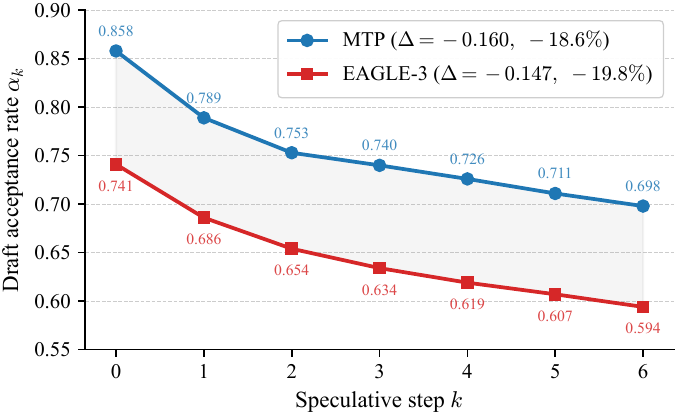}
            \captionsetup{justification=centering,singlelinecheck=true}
            \caption{Step-wise acceptance decay.}
            \label{fig:long-range-decay-curve}
        \end{subfigure}%
    }
    \caption{\textbf{Long-range decay in hidden-state-based drafting.}
    (a)~Later draft steps condition on recursively generated draft hidden
    states, while the corresponding target hidden states are unavailable
    during drafting (dashed gray arrows).  This train-inference mismatch
    accumulates with the speculative step.
    (b)~The effect appears empirically as decreasing draft acceptance rates
    for both a Qwen3.5-4B MTP drafter and EAGLE-3 as the speculative
    step~$k$ increases.}
    \label{fig:long-range-decay}
\end{figure}

Among the various drafter designs, \emph{hidden-state reuse} has become the
dominant paradigm.  EAGLE-style drafters~\citep{eagle,eagle2,eagle3} feed the
target model's internal hidden states into a single-layer drafter to predict
multiple future tokens at low cost.  Multi-token prediction (MTP) drafters
follow a similar reuse strategy and have been adopted in production systems~\citep{medusa,mtp,deepseekv3}.  Despite their practical success, these
hidden-state-based drafters share a common weakness:
\emph{\kvshotblue{long-range decay}}.  Draft accuracy drops
progressively as the speculative step $k$ increases
(~\Cref{fig:long-range-decay}), 
% directly limiting the depth of the draft
% tree and the overall speedup that speculative decoding can deliver.
directly capping the viable depth of the draft tree and the maximum end-to-end speedup that speculative decoding can deliver.

The standard explanation for long-range decay is \emph{train-inference
mismatch}.  During training the draft model conditions on target hidden
states, but at inference it must rely on its own recursively generated
hidden states, which drift further from the target distribution at each
successive step.  Autoregressive \emph{test-time training} (TTT), first
proposed by HASS \citep{hass} and later integrated into EAGLE-3
\citep{eagle3}, addresses this gap by exposing the drafter to its own
drifting trajectories during training.  While TTT measurably improves draft acceptance,~\Cref{fig:long-range-decay-curve} illustrates that long-range decay persists even in TTT-trained drafters, indicating that train-inference mismatch alone cannot fully account for this phenomenon.

In this work, we approach the remaining decay from the perspective of
\emph{context information preservation}. We argue that hidden states act as a
form of \emph{biased context compression}. During hidden-state computation, the
target model's attention mechanism aggregates value vectors according to the
query at the current decoding position, which is optimized for \emph{immediate}
next-token prediction. Tokens that are weakly relevant to this prediction receive
small attention weights, causing their information to be largely suppressed in
the resulting hidden state. This compression is well suited to short-range
prediction, but can weaken the signal needed for later speculative steps. A draft
model that receives only this compressed representation therefore faces an
increasingly difficult information-recovery problem as the prediction horizon
grows.

In contrast, the target model's KV cache serves as an \emph{explicit context
memory} by retaining the complete set of per-position key/value pairs before they
are aggregated by attention. Given these pairs, a draft model can re-attend to
the prefix with its own estimated future queries, making each prefix position
explicitly accessible. The challenge then shifts from recovering suppressed
information to accurately estimating future queries, framing the task as a
function-approximation problem rather than an information-recovery one. This
contrast motivates the \emph{KV-Reuse Hypothesis}: reusing the target model's KV
cache can better preserve context information for long-range drafting, and
therefore \emph{KV-only reuse} should degrade more gracefully than
\emph{hidden-only reuse} at longer speculative steps.

% Cross-attention-style KV reuse has appeared in prior speculative decoding
% systems \citep{glide,longspec}, but it has not been systematically compared
% against hidden-state reuse as a way to understand long-range decay.  To
% fill this gap, we build \textit{KVShot}, a diagnostic framework that
% evaluates KV-based drafting under the same autoregressive TTT pipeline
% used by EAGLE-3.  We organize the comparison around three reuse settings. The first is \emph{hidden-only reuse}, represented by EAGLE-3 and MTP, where target hidden states are concatenated to the drafter input. The second is \emph{KV-only reuse}, where target KV cache is injected into the drafter attention. The third is \emph{hybrid reuse}, where hidden states provide the main anchor and KV reuse contributes through a gated delta correction.
% We compare KV-only and hybrid drafters against a matched EAGLE-3 baseline on Qwen3-8B
% \citep{qwen-3}, isolating the effect of representation choice from other
% confounds.

% zj
Testing this hypothesis requires isolating the effect of the reused
representation itself. Although cross-attention-style KV reuse has appeared in
prior speculative decoding systems \citep{glide,longspec}, it has not been
systematically compared against hidden-state reuse as a mechanism for explaining
long-range decay. To fill this gap, we build \textit{KVShot}, a diagnostic
framework that evaluates KV-Reuse drafting under the same autoregressive TTT
pipeline used by EAGLE-3. We organize the comparison around three reuse settings:
\emph{hidden-only reuse}, represented by EAGLE-3 and MTP, where target hidden
states are concatenated to the drafter input; \emph{KV-only reuse}, where the
target KV cache is injected into drafter attention; and \emph{hybrid reuse},
where hidden states provide the main anchor and KV reuse contributes through a
gated delta correction. We compare KV-only and hybrid drafters against a matched
hidden-only EAGLE-3 baseline on Qwen3-8B \citep{qwen-3}, thereby isolating the
effect of representation choice from other confounds.

Our experiments show that KV reuse degrades more gracefully than hidden-state reuse at longer draft steps. However, these gains remain too small to produce a significant end-to-end speedup. We conduct a detailed analysis of this gap and identify structural limitations of the autoregressive TTT pipeline that prevent it from effectively learning to exploit the KV cache. Our findings suggest that realizing the potential of KV reuse likely requires alternative training paradigms, such as block-wise training \citep{chen2026dflash}, that are better aligned with KV learning.

In summary, this paper makes the following contributions:
\begin{itemize}[leftmargin=19pt]
    \item[(1)] \textit{\textbf{Novel information-preservation view of long-range decay.}}
    We show that long-range decay is not solely a train-inference mismatch
    effect, but also depends on which target-model representation is reused.
    By analyzing hidden states as query-dependent compressed summaries, we
    explain why hidden-only reuse can suppress context needed at later draft
    steps. This leads to a KV-reuse hypothesis: because KV caches preserve
    token-level prefix information before attention aggregation, KV-only reuse
    should degrade more gracefully than hidden-only reuse at longer draft
    horizons (Section~\ref{sec:analysis}).

    \item[(2)] \textit{\textbf{Unified diagnostic framework for representation reuse.}}
    We introduce \textit{KVShot}, a controlled framework that systematically
    compares hidden-only reuse, KV-only reuse, and gated hybrid reuse under
    the same autoregressive TTT pipeline. This design isolates the effect of
    representation choice from other confounds in speculative decoding
    (Section~\ref{sec:method}).

    \item[(3)] \textit{\textbf{Structural bottleneck analysis of KV-Reuse drafting.}}
    We show that KV reuse improves long-range acceptance but remains limited
    under current autoregressive TTT. Our analysis identifies query-estimation
    difficulty and sparse KV-projection gradients as two key bottlenecks,
    motivating training paradigms better aligned with block-level KV learning
    (Section~\ref{sec:discussion}).
\end{itemize}

The paper is structured as follows: \cref{sec:analysis} presents our information-preservation analysis. \cref{sec:method} introduces the KVShot framework and evaluates its empirical performance. \cref{sec:discussion} provides a deeper examination of the bottlenecks associated with autoregressive TTT for KV reuse. We then discuss related work in \cref{sec:related_work}, before concluding in \cref{sec:conclusion}.

\section{Context Information Preservation View of Long-Range Decay}
\label{sec:analysis}

% Section~\ref{sec:intro} frames long-range decay partly as a
% representation-reuse problem. In this section, we make that intuition precise
% by analyzing how hidden-state reuse and KV reuse preserve information
% differently, and derive testable predictions for the experiments that follow.
In this section, we formalize this intuition by comparing hidden-state reuse
and KV reuse in terms of context-information preservation, and derive a
KV-reuse hypothesis together with three testable predictions for the
experiments that follow.

\subsection{Hidden States as Biased Context Compression}
\label{sec:hidden-biased}

To isolate the key mechanism, we focus on the attention aggregation
component of the Transformer block.  A full hidden state additionally
includes output projection, residual connections, layer normalization,
and the feed-forward sub-layer; we omit these for clarity, as the
compression argument applies specifically to the attention step.

Consider a single attention head at layer $\ell$ of the target model.
Given the prefix $x_1, \dots, x_t$, the attention output at position $t$
is a weighted aggregation over value vectors:
\begin{equation}
    h_t^{\ell} \;=\; \sum_{i=1}^{t} \alpha_i \, v_i^{\ell},
    \qquad
    \alpha_i \;=\; \frac{\exp(q_t^{\ell\top} k_i^{\ell})}
                        {\sum_{j=1}^{t} \exp(q_t^{\ell\top} k_j^{\ell})}.
    \label{eq:hidden-state}
\end{equation}
The weights $\{\alpha_i\}$ are determined by the \emph{current} query
$q_t^{\ell}$, which is optimized for predicting $x_{t+1}$.  If a
historical token $x_k$ is weakly relevant to this prediction, the
corresponding weight $\alpha_k$ is near zero and the feature $v_k^{\ell}$
is effectively discarded from $h_t^{\ell}$.

This means $h_t^{\ell}$ is a \emph{query-dependent} compression: it
retains information useful for the immediate next-token prediction but
weakens information that may be critical for predicting tokens further
ahead ($x_{t+2}, x_{t+3}, \dots$).  When a draft model receives
$h_t^{\ell}$ as its input and attempts to predict multiple future tokens,
it faces a difficult recovery problem: the suppressed $v_k^{\ell}$
must be disentangled from a mixture in which they have been weighted
to near zero.  In general, this recovery is inherently ill-conditioned: as attention concentrates more aggressively on a select few positions, it becomes increasingly difficult for a downstream model to reconstruct the weakened components.

This analysis also explains why recent drafters such as EAGLE-3
\citep{eagle3} fuse hidden states from multiple layers rather than
relying on the top layer alone.  Different layers attend to different
aspects of the input, so a token that is suppressed ($\alpha_k \approx 0$)
at one layer may receive substantial weight at another.  Multi-layer
fusion therefore partially recovers information lost by any single
layer's compression.  However, this remains a partial remedy: the
fused representation is still a fixed-dimensional aggregation, and
information that is consistently unattended across all fused layers
remains difficult to recover.

\begin{figure}[t]
    \centering
    \includegraphics[width=0.9\linewidth]{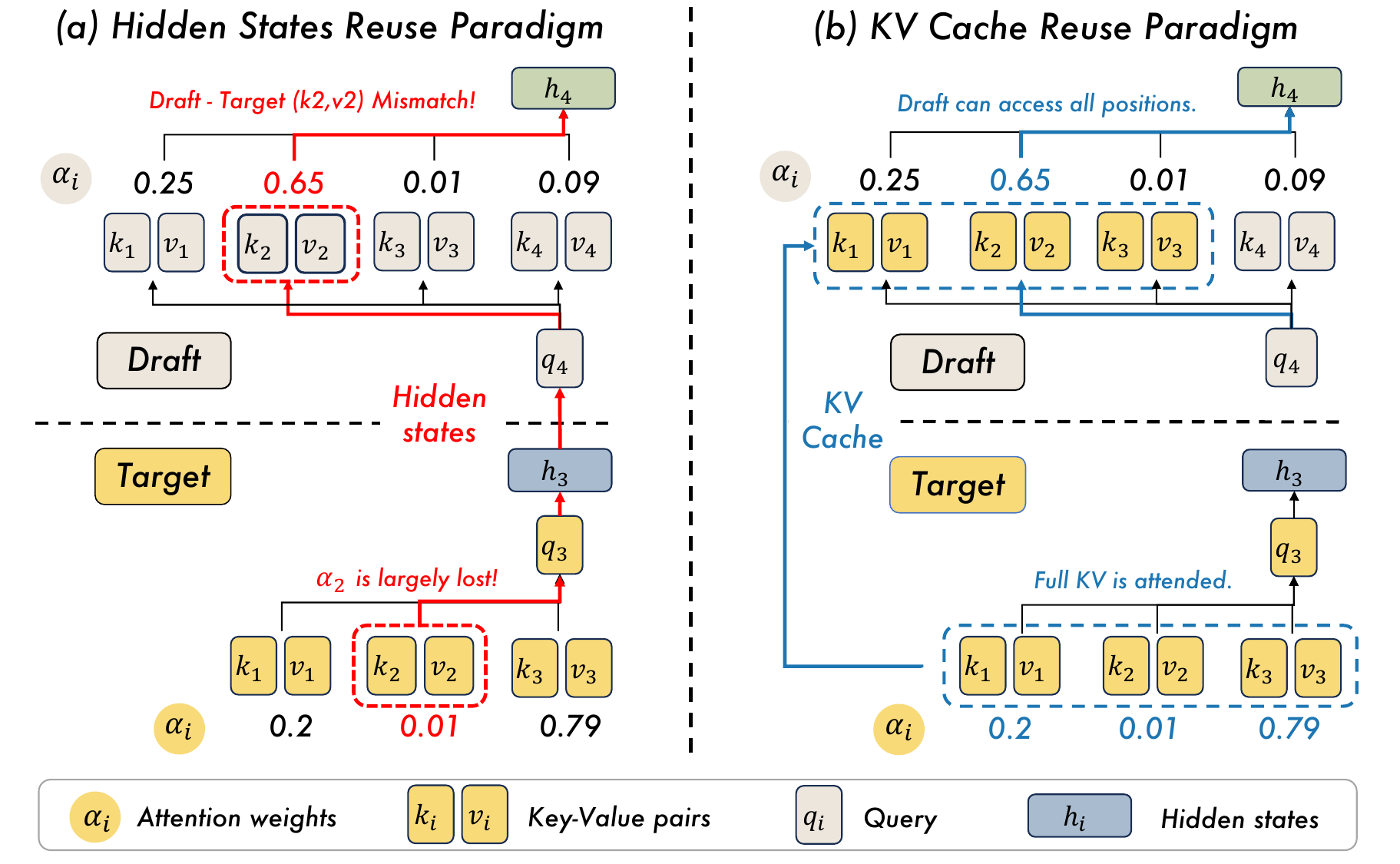}
    \caption{\textbf{Hidden-state reuse vs.\ KV reuse}
    (illustrative; the attention weights shown are schematic, not
    measured).
    (a)~The target model compresses its KV cache into $h_3$ via
    attention; the draft model receives only this aggregated hidden
    state and must predict $x_4$ from it.  Information about
    weakly-attended tokens (e.g.\ $\alpha_2 {=} 0.01$) is largely
    discarded.
    (b)~The target's KV cache is passed directly to the draft model,
    which performs re-attention with its own query~$q_4$.  All
    positions remain accessible, and the approximation error reduces
    to the query estimation error.}
    \label{fig:two-paradigm}
\end{figure}

Figure~\ref{fig:two-paradigm} contrasts the two reuse paradigms.  In the
hidden-state paradigm~(a), token~$x_2$ receives near-zero weight
($\alpha_2 = 0.01$) during the target's aggregation and its information
is largely lost in $h_3$.  Yet the draft model's future prediction
assigns the highest weight to that same token ($\alpha_2 = 0.65$),
illustrating the mismatch between what the target \emph{compressed away}
and what the draft \emph{needs}.  The KV-cache paradigm~(b) avoids this
loss: the draft model re-attends to the full set of target key/value
pairs with its own query, so every position remains accessible.

\subsection{KV Cache as Re-attention Without Prior Aggregation}
\label{sec:kv-reattention}

KV cache preserves the full set of key/value pairs
$\{(k_i^{\ell},\, v_i^{\ell})\}_{i=1}^{t}$ without any lossy
aggregation.  If the draft model is given access to these pairs, it
can produce its own query $q'$ and perform a fresh re-attention:
\begin{equation}
    a' \;=\; \sum_{i=1}^{t} \alpha_i(q') \, v_i^{\ell},
    \qquad
    \alpha_i(q') \;=\;
    \frac{\exp(q'^{\top} k_i^{\ell})}
         {\sum_{j=1}^{t} \exp(q'^{\top} k_j^{\ell})}.
    \label{eq:kv-reattention}
\end{equation}
In the simplified case where the draft query $q'$ closely approximates
the true target query $q_{t+k}^{\ell}$ at some future position $t+k$,
$a'$ approaches the target attention output.  Unlike hidden-state
reuse, no information is discarded by a prior aggregation step.  In
this simplified setting, the approximation quality is governed primarily
by the query estimation error, which we write roughly as
\begin{equation}
    \mathcal{E}_{\mathrm{KV}}
    \;\approx\; \mathcal{E}_{q}
    \qquad \text{(query-error dominated case)}.
    \label{eq:error-kv}
\end{equation}
Equation~\ref{eq:error-kv} is a conceptual breakdown rather than a
formal bound: it isolates the dominant source of error in the
simplified KV-reuse setting and is meant to motivate the experiments
that follow, not to be taken as a tight inequality.
At the same time, a natural objection is that hidden states and KV cache
are still closely related signals rather than fundamentally separate
ones.  We address that objection in Section~\ref{sec:hs-kv-relation}.

\subsection{Are Hidden States and KV Cache Equivalent?}
\label{sec:hs-kv-relation}

A natural objection is that hidden states and KV cache are not
fundamentally different signals.  In a standard Transformer block, the
KV cache at layer $\ell{+}1$ is derived from the hidden state at
layer~$\ell$ via normalization and linear projections:
\begin{equation}
    k_t^{\ell+1} = W_K^{\ell+1}\,\mathrm{Norm}(h_t^{\ell}),
    \qquad
    v_t^{\ell+1} = W_V^{\ell+1}\,\mathrm{Norm}(h_t^{\ell}).
    \label{eq:hs-to-kv}
\end{equation}
This suggests that $\mathrm{KV}_{\ell+1}$ is a simple function of
$h^{\ell}$, seemingly undermining the claim that KV reuse provides
meaningfully different information.  Three observations clarify why
the distinction can still matter in practice.

\paragraph{Last-layer information gap.}
Consider an $L$-layer target model whose top hidden state $h_t^L$ is
reused by the drafter.  The KV cache at layer~$L$ was already consumed by the attention mechanism at layer~$L$ to produce $h_t^L$, a process governed strictly by the biased compression of \autoref{eq:hidden-state}.  There is no layer
$L{+}1$ whose KV cache would correspond to $h_t^L$.  Since prior work
has consistently found that the top layer carries the strongest
predictive signal \citep{medusa,eagle}, the loss of
top-layer KV information may therefore be especially important.

\paragraph{Projection gap.}
Even at intermediate layers, recovering KV from hidden states is not
free.  The projections $W_K$ and $W_V$ are non-trivial transformations;
in models with grouped-query attention, the hidden dimension is typically
$4\sim8{\times}$ larger than the per-head KV dimension.  A draft model that 
receives $h^{\ell}$ instead of the pre-computed $(k^{\ell+1}, v^{\ell+1})$ must 
implicitly learn these projections, imposing an additional burden on a drafter 
with only one or two layers.

\paragraph{Capacity competition.}
Under KV reuse, the drafter's main task is query estimation: producing
$q'$ that approximates the target query.  Under hidden-state reuse,
the same shallow drafter must simultaneously perform query estimation
\emph{and} implicitly reconstruct the KV projections from the received
hidden states.  These two tasks compete for the drafter's limited
representational capacity.  KV reuse removes one of these
responsibilities by providing the pre-computed key/value pairs directly,
concentrating all capacity on query estimation.  Empirically, this
projection gap is observable: directly reusing target KV substantially
outperforms a variant in which the drafter must derive KV from hidden
states via a learned projection (\Cref{app:ablations},
``Hidden$\to$KV cross-projection'').

\subsection{Comparing the Two Error Regimes}
\label{sec:error-comparison}

% Under hidden-state reuse, the draft model's prediction error at step $k$
% can be roughly broken into two interacting components:
% \begin{equation}
%     \mathcal{E}_{\mathrm{HS}}^{(k)}
%     \;\approx\; \underbrace{\mathcal{E}_{\mathrm{comp}}}_{\text{compression loss}}
%     \;+\; \underbrace{\mathcal{E}_{\mathrm{drift}}^{(k)}}_{\text{recursive drift}}.
%     \label{eq:error-hs}
% \end{equation}
% The first term $\mathcal{E}_{\mathrm{comp}}$ reflects the information
% discarded by the target model's attention aggregation: it is present even
% at $k{=}1$ and is independent of the draft model's quality.  The second
% term $\mathcal{E}_{\mathrm{drift}}^{(k)}$ captures the accumulating
% mismatch between target and draft hidden states as the draft model
% recurses over its own outputs; it grows with $k$.  Train-inference
% mismatch via TTT targets $\mathcal{E}_{\mathrm{drift}}$ but cannot
% reduce $\mathcal{E}_{\mathrm{comp}}$, since the compression happens
% inside the target model.

% zj
Under hidden-state reuse, we denote the draft model's prediction error at step $k$
by $\mathcal{E}_{\mathrm{HS}}^{(k)}$, which can be roughly decomposed into two
coupled sources:
\begin{equation}
    \mathcal{E}_{\mathrm{HS}}^{(k)}
    \;\approx\; \underbrace{\mathcal{E}_{\mathrm{comp}}}_{\text{compression loss}}
    \;+\; \underbrace{\mathcal{E}_{\mathrm{drift}}^{(k)}}_{\text{recursive drift}}.
    \label{eq:error-hs}
\end{equation}
The first term $\mathcal{E}_{\mathrm{comp}}$ captures the information loss induced
by the target model's attention aggregation. It is already present at $k{=}1$ and
does not depend on the capacity or quality of the draft model. The second term
$\mathcal{E}_{\mathrm{drift}}^{(k)}$ captures the accumulated mismatch between the
target and draft hidden states as the draft model recursively conditions on its
own predictions, and therefore increases with $k$. These two sources are coupled
because the compressed target hidden states provide the initial condition for
draft recursion, so any information discarded by aggregation can propagate and be
amplified through subsequent drift. Train-inference mismatch via TTT primarily
targets $\mathcal{E}_{\mathrm{drift}}^{(k)}$, but cannot reduce
$\mathcal{E}_{\mathrm{comp}}$, since this compression occurs inside the target
model.

Under KV reuse, the compression component is reduced in the simplified
view behind \autoref{eq:error-kv}, because no prior aggregation discards
information before the draft model sees it.  The dominant challenge then
becomes $\mathcal{E}_{q}$, the draft model's ability to produce an
effective query.  This shifts the difficulty from an
information-recovery problem (disentangling a lossy compression) to a
function-approximation problem (estimating future target queries), while
keeping the underlying key/value pairs available.
The practical takeaway is therefore simple: \textit{KV reuse can preserve more of the
relevant prefix information, but it helps only if the drafter can form useful
future queries from a limited depth and training signal.}

\subsection{KV-Reuse Hypothesis}
\label{sec:predictions}
The preceding analysis suggests a simple hypothesis: because the KV cache
preserves token-level prefix information before attention aggregation, KV reuse
should provide a more robust signal than hidden-state reuse for long-range
speculative drafting. At the same time, exploiting this signal requires the
drafter to estimate useful future queries, which may introduce new bottlenecks.
This hypothesis leads to three testable predictions that guide the experiments
in Sections~\ref{sec:method}--\ref{sec:discussion}:

% This analysis yields three predictions that guide the experiments in
% Sections~\ref{sec:method}--\ref{sec:discussion}:

\begin{itemize}[
    itemsep=1.5pt,
    topsep=1.5pt,
    leftmargin=0pt,
    label={},
    labelsep=0pt,
    itemindent=0pt,
    listparindent=0pt,
    parsep=0pt
]
    \item \textbf{\textsc{Prediction 1} Long-range advantage.}\; KV reuse should degrade more
    gracefully than hidden-state reuse as the speculative step $k$
    increases.  As $k$ grows, the future query $q_{t+k}$ diverges further from the current query $q_t$. Consequently, the hidden-state compression, which is specifically tailored to $q_t$, becomes increasingly prone to discarding information required by $q_{t+k}$.  KV reuse sidesteps
    this mismatch, and its relative advantage should therefore be most
    visible at later steps ($k \geq 3$).

    \item \textbf{\textsc{Prediction 2} Query-estimation bottleneck.}\; The benefit of KV
    reuse depends on how well the draft model can estimate $q'$.
    A very shallow drafter whose queries are linear projections of
    input embeddings will underperform because it lacks the depth to
    approximate target queries, which are produced by many layers of
    nonlinear transformation.

    \item \textbf{\textsc{Prediction 3} Short-range disadvantage.}\; At $k{=}0$ or $k{=}1$,
    hidden-state reuse may still be preferable: $\mathcal{E}_{\mathrm{comp}}$
    is small when the information relevant to the next token is also the
    most attended, and hidden states carry richer semantic content than
    raw input embeddings fed to a KV-based drafter.
\end{itemize}

If all three predictions hold, the practical implication is that KV
reuse is most valuable when combined with hidden-state reuse in a hybrid
design, and that the training pipeline must provide enough capacity
and a gradient signal for the draft model to learn effective queries.

\begin{figure}[t]
    \centering
    \includegraphics[width=0.9\linewidth]{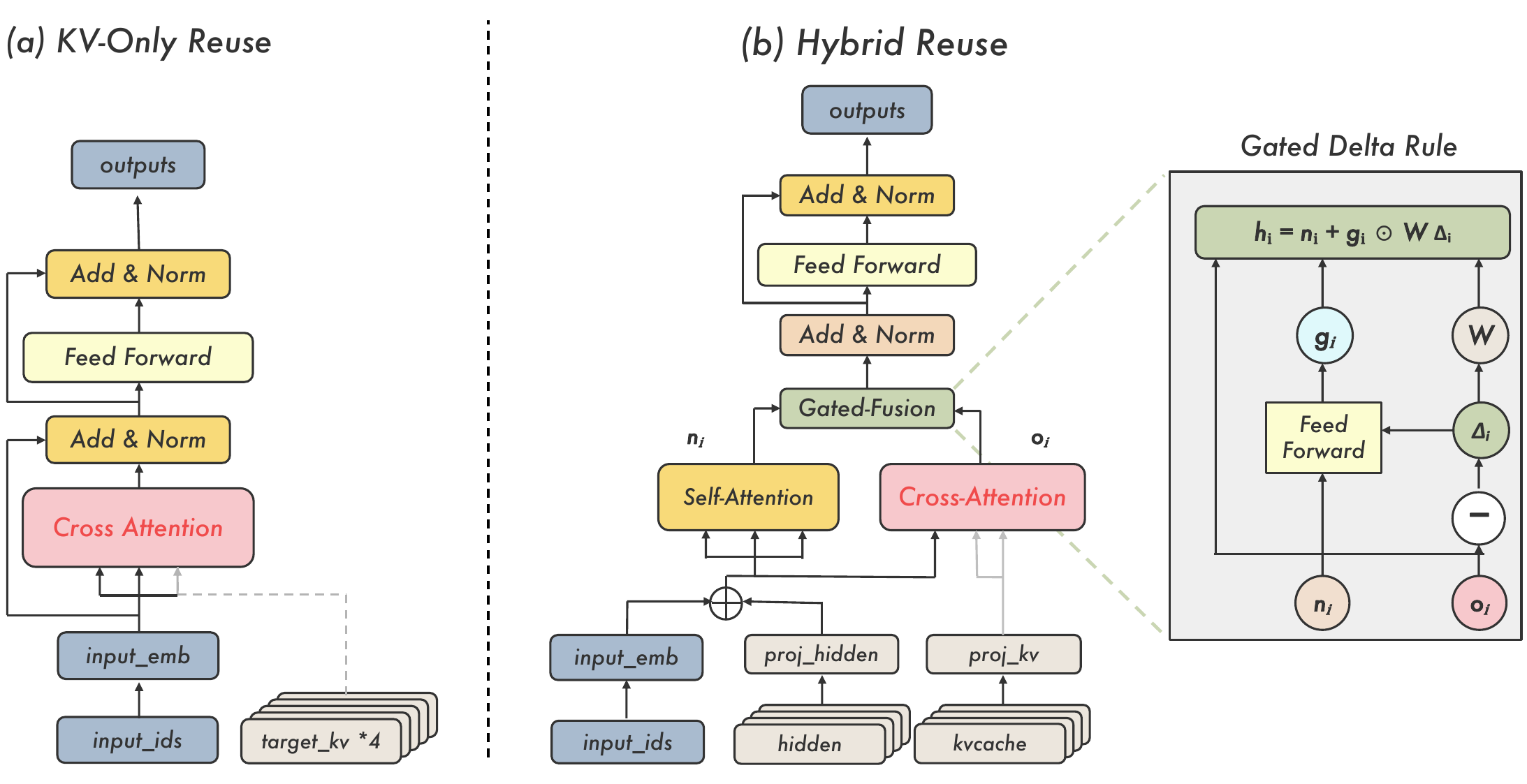}
    \caption{\textbf{KV Reuse architectures in \textit{KVShot}.}
    (a) \textbf{KV-only reuse} injects the target model's KV cache into the drafter
through cross-attention, allowing the drafter to directly attend to token-wise
prefix representations. (b) \textbf{Hybrid reuse} combines hidden-state reuse
with KV reuse: the hidden pathway provides the main draft representation, while
the KV pathway supplies a gated delta correction through cross-attention. The
gated delta rule adaptively controls how much KV-derived information is fused
into the hidden representation.}
    \label{fig:framwork}
\end{figure}

\section{KVShot: Testing the KV-Reuse Hypothesis}
\label{sec:method}

This section reports the experimental study built on the
\textit{KVShot} framework, as illustrated in ~\Cref{fig:framwork}.  Throughout this section, \textit{KVShot}
refers to the overall diagnostic setup rather than to a single model;
we refer to the specific model families as \emph{KV-only drafters} and
\emph{hybrid drafters}.

% ------------------------------------------------------------------
\subsection{Experimental Setup}
\label{sec:setup}

\paragraph{Target model and data.}
All experiments use Qwen3-8B \citep{qwen-3} as the target
model.  Initial ablations (Sections~\ref{sec:pure-kv}--\ref{sec:hybrid})
train on ShareGPT~\citep{sharegpt-vicuna-unfiltered}
(${\sim}$70k samples, 3~epochs) for fast iteration.
The end-to-end evaluation in~\cref{sec:e2e} scales up to 280k
samples (ShareGPT + UltraChat) with target-model--regenerated responses.

\paragraph{Training.}
We implement the experiments on top of SpecForge~\citep{specforge2025} and train all drafters with the autoregressive test-time training (TTT) objective used by EAGLE-3
\citep{eagle3}, which aligns training with inference-time hidden-state
drift.

\paragraph{Metrics.}
We report step-wise draft acceptance rates $\alpha_k$
($k = 0, \dots, 6$) and the expected mean accepted tokens
$\mathrm{MAT} = 1 + \sum_{k=0}^{K-1}\prod_{j=0}^{k}\alpha_j$.
For the end-to-end evaluation, we additionally report the MAT measured
with the HuggingFace speculative decoding pipeline using a draft tree
configuration of $(8, 10, 60)$.

\paragraph{Baseline.}
The primary baseline is a 1-layer EAGLE-3 drafter trained under the
same setting as each KV-based variant.

% ------------------------------------------------------------------
\subsection{KV-Only Reuse: Does Re-attention Help?}
\label{sec:pure-kv}

\paragraph{Design.}
As shown in ~\Cref{fig:framwork} (a), we replace the hidden-state input entirely and inject the target KV cache into the draft model's attention.  We compare two injection strategies for
handling multi-layer target KV:
\ding{172} \emph{head concatenation}, which expands draft KV heads by a factor of
$L_s$ and concatenates the KV from $L_s$ sampled target layers; and
\ding{173} \emph{linear projection}, which maps the concatenated multi-layer KV
back to a single-layer representation via learned matrices
$W_K^{\mathrm{proj}}, W_V^{\mathrm{proj}} \in
\mathbb{R}^{d_{\mathrm{kv}} \times L_s \cdot d_{\mathrm{kv}}}$.
The linear projection destroys positional information already encoded
via RoPE; \ding{174} we test a variant that re-applies RoPE after projection.

\paragraph{Results.}

\begin{table}[t]
\centering
\caption{KV-only reuse ablations (1-layer drafter, ShareGPT training).
    All rows except the EAGLE-3 baseline receive \emph{no} target hidden
    states.  $\alpha_k$ denotes the draft acceptance rate at step~$k$.
    `Retention' denotes the long-range retention ratio
    $\alpha_6 / \alpha_0$.}
\label{tab:pure-kv}
\small
\begin{tabular}{@{}lccccc@{}}
\toprule
Method & $\alpha_0$ & $\alpha_3$ & $\alpha_6$ & Retention & MAT \\
\midrule
No target info   & 0.237 & 0.237 & 0.236 & 99.6\% & 1.31 \\
Head concat      & 0.489 & 0.332 & 0.305 & 62.4\% & 1.78 \\
Linear projection& 0.488 & 0.379 & 0.354 & 72.5\% & 1.83 \\
Proj + RoPE fix  & 0.494 & 0.381 & 0.353 & 71.5\% & 1.84 \\
\midrule
EAGLE-3 (baseline) & 0.638 & 0.511 & 0.469 & 73.5\% & 2.37 \\
\bottomrule
\end{tabular}
\end{table}

\Cref{tab:pure-kv} shows the results.  Three findings stand out.
(1)~Target KV provides a clear signal: removing it entirely (``No target
info'') drops MAT from 1.84 to 1.31.
(2)~Linear projection substantially outperforms head
concatenation (MAT 1.83 vs.\ 1.78), suggesting that forcing a single
query to attend across $L_s$ disjoint KV spaces is too difficult for
a single-layer drafter.
(3)~Re-applying RoPE after projection yields only a marginal
gain (1.84 vs.\ 1.83), suggesting that positional corruption is not
the dominant performance bottleneck.

Despite these improvements, all KV-only variants remain far below the
EAGLE-3 baseline (MAT 2.37).  In particular, $\alpha_0$ never exceeds
0.50, versus 0.64 for EAGLE-3.  This is consistent with \textsc{Prediction~3} from
Section~\ref{sec:predictions}: at short range, hidden states carry
richer semantic content than the raw input embeddings that a
KV-only drafter must rely on.  At the same time, the gap between
``No target info'' and the KV-reuse variants show that target KV does
matter, which motivates the next test of \textsc{Prediction~2}: whether better
query estimation can unlock more of that signal.
The retention column should therefore be read together with the
absolute acceptance rates: the nearly flat but uniformly low
``No target info'' curve yields a nominal 99.6\% retention, yet still
performs much worse overall.

% ------------------------------------------------------------------
\subsection{Depth Scaling: Is Query Estimation the Bottleneck?}
\label{sec:depth-scaling}

\paragraph{Design.}
This subsection directly tests \textsc{Prediction~2} from
Section~\ref{sec:predictions}: KV reuse should help only if the draft
model can estimate useful queries.  A 1-layer drafter's queries are
linear projections of input embeddings and lack the depth to approximate
target queries.  We therefore progressively increase the drafter from 1
to 4 layers, all using the linear-projection injection with RoPE
re-application.

\paragraph{Results.}

\begin{table}[t]
\centering
\caption{Effect of drafter depth on KV-only reuse (ShareGPT training).
    `Retention' denotes the long-range retention ratio
    $\alpha_6 / \alpha_0$.  The EAGLE-3 baseline uses 1 layer with
    hidden-state reuse.}
\label{tab:depth}
\small
\begin{tabular}{@{}lcccccc@{}}
\toprule
Drafter & Layers & $\alpha_0$ & $\alpha_3$ & $\alpha_6$ & Retention & MAT \\
\midrule
KV-only & 1 & 0.494 & 0.381 & 0.353 & 71.5\% & 1.84 \\
KV-only & 2 & 0.594 & 0.497 & 0.463 & 77.9\% & 2.23 \\
KV-only & 3 & 0.609 & 0.518 & 0.487 & 80.0\% & 2.31 \\
KV-only & 4 & 0.614 & 0.527 & 0.495 & 80.6\% & 2.34 \\
\midrule
EAGLE-3 & 1 & 0.638 & 0.511 & 0.469 & 73.5\% & 2.37 \\
\bottomrule
\end{tabular}
\end{table}

\Cref{tab:depth} supports query estimation as a major bottleneck.
Moving from 1 to 2 layers produces the largest single gain
($\Delta\mathrm{MAT} = {+}0.39$), because the second layer gives the
drafter a context-aware query rather than a linear transform of the
input embedding.  Returns diminish rapidly: 3 and 4 layers add only
$+0.08$ and $+0.03$ respectively.  Even a 4-layer KV-only drafter
(MAT 2.34) only approaches, but does not exceed, the 1-layer EAGLE-3
baseline (MAT 2.37)---at significantly higher drafting cost.  This
diminishing-return pattern could partly reflect optimization difficulty
in deeper KV-only drafters rather than a pure capacity limit.  Still, the
sharp jump from 1 to 2 layers strongly suggests that giving the drafter
enough depth to form a context-aware query is a major part of the
bottleneck.

The retention column makes the long-range pattern more explicit.
As depth increases, the KV-only retention ratio $\alpha_6/\alpha_0$
rises from 71.5\% to 80.6\%, eventually exceeding the EAGLE-3 baseline
(73.5\%).  This is exactly the behavior anticipated by Prediction~1:
once query estimation is strong enough, KV reuse degrades less
severely across speculative steps even if its short-range accuracy
still trails hidden-state reuse.

Notably, at $k{=}6$, the 4-layer KV-only drafter (0.495) does surpass
EAGLE-3 (0.469), supporting the long-range advantage anticipated by
Prediction~1, while at $k{=}0$ it remains below EAGLE-3
(0.614 vs.\ 0.638), as anticipated by Prediction~3.  These two effects
roughly cancel in the aggregate MAT.  In other words, the depth-scaling
study supports all three predictions together: better query estimation
helps, KV reuse becomes more competitive at long range, and hidden-state
reuse still retains the short-range edge.

% ------------------------------------------------------------------
\subsection{Hybrid Reuse: Can KV Complement Hidden States?}
\label{sec:hybrid}

The KV-only experiments now resolve the prediction pattern from
Section~\ref{sec:predictions} more clearly.  \textsc{Prediction~1} appears at
long range, where KV reuse becomes more competitive.  Prediction~3
appears at short range, where hidden-state reuse remains stronger.
If both effects are real, the natural next step is a hybrid drafter
that keeps the hidden-state anchor for early steps and lets KV reuse
provide a correction at a longer range.

\subsubsection{Gated Delta Rule}

As shown in ~\Cref{fig:framwork} (b), we augment the EAGLE-3 self-attention path with a parallel cross-attention
path to target KV, merged via a \emph{gated delta} rule.  The design
intuition is simple: the hidden-state path should preserve the strong
short-range behavior of EAGLE-3, while the KV path should contribute a
correction when longer-range re-attention becomes useful.
Given the draft token representation $x_i$, we compute:
\begin{align}
    n_i &= \mathrm{SelfAttn}(x_i),
        &\text{(EAGLE-3 path)} \label{eq:self-attn} \\
    o_i &= \mathrm{CrossAttn}(x_i,\;\mathcal{M}_i),
        &\text{(KV path)} \label{eq:cross-attn}
\end{align}
where $\mathcal{M}_i$ contains the target KV cache for verified prefix
tokens and draft-generated KV for subsequent positions.  The
cross-attention output is treated as a correction:
\begin{equation}
    \Delta_i = o_i - n_i.
    \label{eq:delta}
\end{equation}
A shared gate controls how much of this correction is applied:
\begin{equation}
    g_i = \sigma\!\left(
        W_g
        \begin{bmatrix} n_i \\ o_i \\ \Delta_i \end{bmatrix}
        + b_g
    \right),
    \qquad
    h_i = n_i + g_i \odot W_{\Delta}\,\Delta_i,
    \label{eq:fusion}
\end{equation}
where $\sigma$ is the sigmoid function and $W_g, b_g$ are shared across
all draft steps.  When $g_i \approx 0$, the model reduces to standard
EAGLE-3; when $g_i$ is large, the cross-attention correction shifts the
representation toward what re-attention to the target KV suggests.  This
design anchors short-range prediction on the self-attention branch while
allowing the cross-attention branch to act as a long-range correction
term.

The model can be trained from random initialization or warm-started from
an existing EAGLE-3 checkpoint.  In the latter case, the self-attention
branch and all shared modules (input projection, MLP, layer norms)
inherit pre-trained weights, while the newly added cross-attention
projections, the gate, and the delta projection are randomly initialized.
The \emph{Cross-only} variant follows the same initialization scheme but
removes the self-attention output from the forward pass entirely; it
still loads the EAGLE-3 query/key/value/output projections and the MLP,
so the only architectural difference from the hybrid drafter is
that $n_i$ is omitted when forming $h_i$.

\subsubsection{Results}

\begin{table}[t]
\centering
\caption{Hybrid drafter results (1-layer drafter, ShareGPT training).
    ``Cross-only'' removes the self-attention output from the forward
    pass; the underlying projections are still inherited from the
    EAGLE-3 checkpoint when applicable.  `Retention' denotes the
    long-range retention ratio $\alpha_6 / \alpha_0$.}
\label{tab:hybrid}
\small
\begin{tabular}{@{}llccccc@{}}
\toprule
Method & Init & $\alpha_0$ & $\alpha_3$ & $\alpha_6$ & Retention & MAT \\
\midrule
EAGLE-3      & --              & 0.638 & 0.511 & 0.469 & 73.5\% & 2.37 \\
\midrule
Hybrid     & random          & 0.650 & 0.527 & 0.490 & 75.4\% & 2.44 \\
Hybrid     & EAGLE-3 ckpt    & 0.665 & 0.553 & 0.514 & 77.3\% & 2.54 \\
\midrule
Cross-only   & random          & 0.634 & 0.499 & 0.450 & 71.0\% & 2.34 \\
Cross-only   & EAGLE-3 ckpt    & 0.637 & 0.495 & 0.442 & 69.4\% & 2.35 \\
\bottomrule
\end{tabular}
\end{table}

\Cref{tab:hybrid} provides the first clear positive step-wise
result in this study.
The hybrid drafter warm-started from an EAGLE-3 checkpoint reaches
MAT~2.54, exceeding the EAGLE-3 baseline (2.37) by $+0.17$.  Improvements
are visible across all steps: $\alpha_0$ rises from 0.638 to 0.665
($+4.2\%$), and $\alpha_6$ from 0.469 to 0.514 ($+9.6\%$), suggesting
that the KV correction is especially helpful at longer range.
The retention ratio makes the same pattern explicit: it rises from
73.5\% for EAGLE-3 to 77.3\% for the checkpoint-initialized hybrid drafter.

Two additional findings are notable.
(1)~Checkpoint initialization matters: warm-starting from an
EAGLE-3 checkpoint (MAT 2.54) significantly outperforms random
initialization (2.44), suggesting that a strong self-attention anchor
accelerates the learning of the cross-attention correction.
(2)~The self-attention anchor is essential: removing it
(``Cross-only'') drops MAT to 2.34--2.35, \emph{below} the EAGLE-3
baseline, even though the underlying projections are still inherited
from the EAGLE-3 checkpoint.  This suggests that cross-attention alone
cannot replace the hidden-state pathway, and supports the gated delta
design in which KV information contributes as a correction rather than
a replacement.  This is exactly the hybrid implication suggested by the
three predictions in~\Cref{sec:predictions}: KV reuse helps most
when it complements hidden-state reuse instead of trying to replace it.

% ------------------------------------------------------------------
\subsection{End-to-End Evaluation: Does the Gain Survive Overhead?}
\label{sec:e2e}

The step-wise acceptance improvements in~\Cref{tab:hybrid} suggest
that the predictions from~\Cref{sec:predictions} are broadly
correct at the per-step level.  The practical question, however, is
whether those validated step-wise effects survive once training is
scaled up and drafting overhead is included in an end-to-end proxy.
We therefore scale up training in two ways.  The dataset grows from
${\sim}$70k to 280k samples (ShareGPT~+~UltraChat), and target
responses are regenerated by Qwen3-8B itself so that the training
distribution matches the inference distribution.  Evaluation uses the
HuggingFace speculative decoding pipeline with a draft tree of shape
$(8, 10, 60)$ (depth~8, top-$k$ at each layer~10, total $\leq 60$
candidate tokens).  In this section, we treat HF-measured MAT as our
primary end-to-end proxy rather than as a direct wall-clock throughput
measurement.

\begin{table}[t]
\centering
\caption{End-to-end evaluation (Qwen3-8B, 280k training samples,
    HF-measured MAT).  The hybrid drafter uses the gated delta rule.}
\label{tab:e2e}
\small
\begin{tabular}{@{}lc@{}}
\toprule
Setting & HF MAT \\
\midrule
EAGLE-3 existing ckpt   & 4.43 \\
EAGLE-3 train from scratch & 4.76 \\
EAGLE-3 train from ckpt & 5.01 \\
\midrule
Hybrid train from ckpt  & 5.04 \\
\bottomrule
\end{tabular}
\end{table}

\Cref{tab:e2e} gives the central negative result of this study.
In the controlled comparison (both trained from the same checkpoint),
the hybrid drafter increases HF-measured MAT only from 5.01 to
5.04 ($+0.6\%$).  Our profiling further indicates that the additional
cross-attention path introduces roughly 5--10\% extra drafting latency.
Taken together, these numbers do not support a meaningful end-to-end
speedup claim in the current pipeline, and suggest that any wall-clock
gain would be marginal at best.

The KV gain shrinks from $+0.17$ MAT in the step-wise evaluation to $+0.03$ HF MAT in the end-to-end proxy, likely because several factors compound in the end-to-end setting. First, the EAGLE-3 baseline
itself benefits substantially from the larger, regenerated dataset
(MAT rising from $4.43$ on the existing checkpoint to $5.01$ when
retrained).  This suggests that part of what the gated KV path was
correcting at a small scale is also recoverable by giving the
hidden-state baseline more and better-aligned training data.  Second,
HF MAT is computed under tree verification with $(8, 10, 60)$, which
already exploits multiple candidate paths and tends to compress the
visible difference between drafters whose step-wise curves differ
mildly.  We do not view either factor as fully explaining the
shrinkage, but together they make a small end-to-end gap consistent
with a non-trivial step-wise gap.

This result does \emph{not} invalidate the information-preservation
analysis of~\Cref{sec:analysis}: the step-wise improvements
in~\Cref{tab:hybrid} show that adding a KV correction path can help
long-range prediction.  Rather, it indicates that the current
autoregressive TTT pipeline cannot exploit this signal efficiently
enough to overcome the added cost.  In other words, validating
Predictions~1--3 are not yet sufficient for an end-to-end speedup claim.
The next section analyzes the deeper pipeline-level causes of this
shrinkage.

\section{Why Does Current Autoregressive TTT Fit KV Reuse Poorly?}
\label{sec:discussion}

The experiments in Section~\ref{sec:method} largely validate the
step-wise predictions from Section~\ref{sec:predictions}.  KV-only
reuse becomes relatively more competitive at later draft steps
(\textsc{Prediction~1}), deeper KV-only drafters improve substantially
(\textsc{Prediction~2}), and hidden-state reuse retains the short-range edge
(\textsc{Prediction~3}).  The remaining question is why these step-wise effects
still collapse into only a marginal end-to-end gain under the current
pipeline.  In this section, we analyze three pipeline-level bottlenecks
that together account for that gap, and discuss what they imply for
future work.

% ------------------------------------------------------------------
\subsection{Query Estimation Is Harder Than It Appears}
\label{sec:query-bottleneck}

The information-preservation analysis in Section~\ref{sec:analysis}
isolates query estimation, denoted by $\mathcal{E}_q$, as the main remaining difficulty for KV reuse. In principle this is a pure function-approximation problem.
In practice, however, the target model's queries at position $t{+}k$
are the output of $L$ layers of nonlinear transformation over the full
prefix.  A shallow draft model with one or two layers has fundamentally
limited capacity to approximate these queries.

The depth-scaling experiments in Section~\ref{sec:depth-scaling}
provide direct evidence: moving from 1 to 2 layers produces a large
MAT jump ($+0.39$), but even 4 layers cannot match a 1-layer EAGLE-3
baseline.  This means that accurate query estimation requires
significantly more model capacity than accurate next-hidden-state
prediction, because the drafter must replicate the multi-layer
compositional structure that produces target queries, rather than
predicting one layer's output from the previous layer's hidden state.

% ------------------------------------------------------------------
\subsection{Sparse Optimization of Draft-Side KV Projections}
\label{sec:kv-gradient}

A second, less obvious bottleneck lies in the training dynamics of
the draft model's own KV projections ($W_K^{\mathrm{cross}}$ and
$W_V^{\mathrm{cross}}$ in \autoref{eq:cross-attn}).  In plain terms,
most of the KV cache seen during training is copied directly from the
target model, so the draft-side KV pathway is updated by only a small
number of draft tokens.

Under autoregressive TTT, each training step processes a sequence of
$K$ draft tokens.  For the cross-attention branch, only the
\emph{draft-generated} portion of the KV cache exercises
$W_K^{\mathrm{cross}}$ and $W_V^{\mathrm{cross}}$.  The prefix portion
is copied from the target and does not backpropagate through these
parameters.  In practice, the number of draft tokens ($K$, typically
$\leq 7$) is tiny relative to the full prefix length.  As a result,
only a small fraction of the attention computation contributes
gradient to the draft-side KV projections.  The optimization is
therefore sparse and unbalanced: the query projection $W_Q$ receives
dense gradient from every position, whereas the KV projections are
updated only by the few draft positions.

We attempted to mitigate this by scaling the KV-projection gradient
by $50\times$, but observed no meaningful improvement
(MAT 2.21 vs.\ 2.18 for the same setting without scaling;
see Appendix~\ref{app:ablations} for full results).
This suggests that the problem is not the gradient magnitude per se
but the diversity of training signal: the same few draft tokens
patterns are repeatedly used to train the KV pathway, preventing it
from generalizing.

% ------------------------------------------------------------------
\subsection{Gate-Induced Gradient Starvation}
\label{sec:gate-starvation}

\begin{figure}[t]
    \centering
    \includegraphics[width=0.9\linewidth]{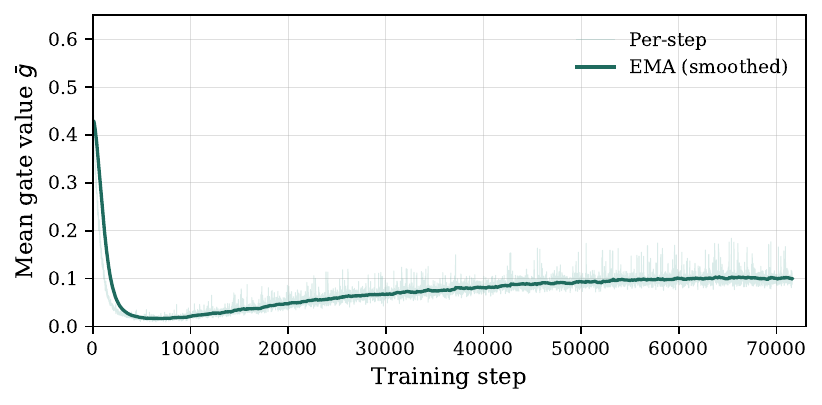}
    \caption{\textbf{Gate-induced gradient starvation.}
    Mean gate value $\bar{g}$ of the warm-started hybrid drafter
    (EAGLE-3 ckpt, ShareGPT, 70k samples) over training.  From the
    initial value $\bar{g}\!\approx\!0.5$ (sigmoid + random $W_g, b_g$),
    the gate collapses to $\bar{g}\!\approx\!0.02$ within the first
    few thousand steps and then slowly recovers to
    $\bar{g}\!\approx\!0.10$ over the remaining $\sim$60k steps.
    The light line is per-step, the dark line is an EMA
    ($\alpha{=}0.02$).}
    \label{fig:gate-trajectory}
\end{figure}

A third, distinct bottleneck appears in the training dynamics of the
hybrid drafter itself.  Recall that the fusion output
(\autoref{eq:fusion}) adds a gated correction $g_i \odot W_{\Delta}\Delta_i$
to the self-attention output $n_i$.  In the ``EAGLE-3 ckpt'' variant,
the self-attention branch is warm-started with strong pretrained
weights, while the cross-attention projections, the gate, and
$W_{\Delta}$ are randomly initialized.  In this configuration, we
consistently observe a characteristic trajectory of the mean gate
value $\bar{g} = \mathbb{E}[\,\mathbb{1}^{\top} g_i / d\,]$
over training, shown in Figure~\ref{fig:gate-trajectory}.

The trajectory is consistent with a gated-residual optimization
failure mode.  At initialization the cross-attention branch is random, so
its output is near-noise relative to the pretrained $n_i$.  The
fastest local direction for reducing the loss is therefore to push
$\bar{g}$ toward zero, effectively silencing the cross-attention
branch and recovering the EAGLE-3 baseline.  Once the gate is closed,
however, the gradient flowing through the cross-attention branch is
itself scaled by $g_i$, so the randomly initialized projections
receive only a weak training signal.  This creates a self-reinforcing
low-injection state.  The gate does not reopen until cross-attention
learns something useful, and cross-attention struggles to learn while
the gate stays small.  The slow, noisy recovery from $0.02$ to $0.10$
visible in Figure~\ref{fig:gate-trajectory} is what this starvation
looks like in practice.

Two qualifications are worth stating explicitly.
First, a small $\bar{g}$ is not the same as ``the cross-attention
branch is ignored'': $\bar{g}$ averages over element-wise gates, and
the positive MAT gap of $+0.17$ over EAGLE-3 in
Table~\ref{tab:hybrid} shows that even $\bar{g}\!\approx\!0.1$ is
enough to inject a useful correction at the ShareGPT-70k scale.
Second, gate starvation on its own does not fully explain the small
end-to-end gap in Table~\ref{tab:e2e}, which is also shaped by the
stronger EAGLE-3 baseline under regenerated 280k data and by tree
verification compressing visible step-wise differences.  Nevertheless, the trajectory identifies a third pipeline-level bottleneck specific to architectures that fuse a warm-started and a randomly initialized branch through a learned gate, a limitation that cannot be resolved by improving query estimation or densifying the KV-projection gradient.

% ------------------------------------------------------------------
\subsection{Implications for Future Training Pipelines}
\label{sec:future}

All three bottlenecks ultimately trace back to the same root cause.
The autoregressive TTT pipeline generates draft tokens one at a time.
That sequential structure limits the depth of query computation and restricts the number of tokens available to train the draft-side KV pathway. Moreover, in gated fusion architectures, it severely weakens the gradient received by the cross-attention branch during the critical early phase of training.

Recent work on non-autoregressive drafting suggests a potential
resolution.  DFlash \citep{chen2026dflash} replaces autoregressive
drafting with a block diffusion adapter that predicts an entire block
of tokens in parallel.  These design choices are consistent with
addressing at least the first two bottlenecks above, though we have not
verified that hypothesis through a controlled comparison here:
\begin{itemize}
    \item A deeper draft model (5~layers in DFlash) provides the
    capacity for more accurate query estimation, while mask-token
    parallelism amortizes the additional inference cost.
    \item Block-wise training generates many draft tokens per step,
    so the KV projections receive a dense gradient from a diverse set
    of positions rather than the sparse signal of autoregressive TTT.
\end{itemize}

Our findings are consistent with this possibility.  The diagnostic value of
KVShot is precisely in pinpointing \emph{where} the current pipeline
falls short.  The problem is not that the predictions in
Section~\ref{sec:predictions} fails at the step-wise level; those
predictions are mostly supported.  The problem is that the training
setup cannot turn that step-wise advantage into an efficient
end-to-end system.  We therefore view block-wise training pipelines as
a promising next direction for making KV reuse a practical
option in speculative decoding.

\section{Related Work}
\label{sec:related_work}

\paragraph{Speculative decoding.}
Speculative decoding was introduced concurrently as a lossless acceleration
strategy that uses a lightweight draft process to propose multiple future
tokens for parallel verification by a target model
\citep{sps1,sps2}. Subsequent systems
work substantially expanded this design space. SpecInfer organizes draft
candidates into token trees for parallel verification
\citep{specinfer}, and Sequoia improves scalability and hardware
adaptation through tree optimization and hardware-aware scheduling
\citep{sequoia}. Draft \& Verify studies self-speculative decoding,
where the target model drafts and verifies using its own intermediate
layers \citep{Zhang_2024}. SWIFT extends this line to on-the-fly
self-speculative decoding \citep{xia2025swiftontheflyselfspeculativedecoding},
and KNN-SSD further studies dynamic self-speculative decoding through
nearest-neighbor layer set optimization \citep{song-etal-2026-knn}.
PEARL studies adaptive draft-length control
\citep{pearl}, TALON builds confidence-aware token trees
\citep{talon}, and SpecBranch introduces rollback-aware branch
parallelism for hybrid drafting \citep{specbranch}. Retrieval-based
variants such as LogitSpec and Double further broaden the design space
by using retrieved candidates or double retrieval parallelism to
accelerate speculation \citep{logitspec,double}. HIPPO further extends
parallel speculative decoding to video large language models through a
holistic-aware design \citep{lv2026hippoacceleratingvideolarge}.
Parallel exact alternatives such as Lookahead Decoding remove the auxiliary
draft model altogether, trading additional computation per step for fewer
sequential decoding steps \citep{lade}. A recent survey summarizes this broader landscape
\citep{xia2024survey}.

\paragraph{Drafters that reuse target hidden states.}
A second line of work focuses on the drafter itself. Medusa augments the
target model with multiple decoding heads to predict several future tokens
in parallel \citep{medusa}, and multi-token prediction (MTP) adopts a
similar objective at training time and is now widely deployed as an
inference-time drafter in production LLMs
\citep{mtp,deepseekv3}. EAGLE \citep{eagle} shifts drafting
from token prediction to hidden-state prediction, showing that reusing
internal target-model features can substantially reduce drafting cost, and
EAGLE-2 extends this line with dynamic draft trees \citep{eagle2}.
HASS addresses the resulting train-decode inconsistency through harmonized
objectives and context alignment \citep{hass}, and EAGLE-3 further
integrates test-time training into the drafter and fuses multi-layer
features \citep{eagle3}.

\paragraph{Drafters that reuse target KV cache.}
A smaller line of work has the drafter consume the target's KV cache
directly through a cross-attention sub-layer.  GLIDE+CAPE \citep{glide}
inserts cross-attention into a shallow drafter so that draft queries
attend to the target's top-layer KV from the previous verification block,
and trains the drafter from scratch with standard teacher-forced
cross-entropy.  LongSpec \citep{longspec} adopts a similar
cross-attention-to-target-KV design, motivated by long-context inference:
because the target's KV is stored anyway, a draft that re-attends to it
keeps drafter memory bounded as context grows, and is paired with
anchor-offset position indices and flash noisy training to handle long
prefixes.  These works establish that cross-attention-style KV reuse is
a practical drafter design.  Our work differs from both in three ways:
(i)~the focus is on long-range decay and information preservation rather
than low-overhead drafting (GLIDE) or long-context memory efficiency
(LongSpec); (ii)~we run a controlled head-to-head between KV reuse and
hidden-state reuse under a single training setup, including hybrid
designs that combine both signals, rather than taking KV reuse as a
fixed design choice; and (iii)~we evaluate KV-based drafters under the
autoregressive TTT objective used by EAGLE-3, rather than standard
teacher-forced training, which is what reveals the
training-pipeline-level bottlenecks discussed in
Section~\ref{sec:discussion}.

\paragraph{Beyond autoregressive drafting.}
Our analysis of why KV reuse struggles under autoregressive TTT is related
to recent non-autoregressive and block-wise drafting directions. DFlash
\citep{chen2026dflash} replaces autoregressive drafting with a block
diffusion adapter conditioned on target hidden features, and argues that
parallel block drafting bypasses the gradient-sparsity problems of
autoregressive TTT pipelines. Our empirical findings are consistent with
this view: they indicate that the main bottleneck for KV reuse lies at
least as much in the training setup as in the choice of reused
representation.

\section{Conclusion}
\label{sec:conclusion}

This paper investigates whether reusing the target model's KV cache
can alleviate long-range decay in speculative decoding better than
the prevailing hidden-state reuse paradigm.  We approach the question
from two directions: an context information preservation analysis that offers a
conceptual breakdown of the two reuse approaches, and a systematic
experimental study on Qwen3-8B using the KVShot framework.

Our analysis suggests that hidden-state reuse can be viewed as
introducing a compression loss---information weakened by the target
model's own attention is difficult for a downstream drafter to
recover---whereas KV reuse can preserve access to the full set of
key/value pairs through re-attention.  In the simplified KV-reuse view,
the dominant difficulty shifts toward query estimation, a
function approximation problem rather than an information-recovery
problem.

The experiments support this picture in a nuanced way.  KV-only
drafters become relatively more competitive at longer draft steps, and a
hybrid drafter that combines KV and hidden-state signals improves
draft acceptance over an EAGLE-3 baseline (MAT 2.54 vs.\ 2.37).
However, the end-to-end proxy remains small: HF-measured MAT rises
only from 5.01 to 5.04, while drafting latency increases by 5--10\%.

We trace this gap to three bottlenecks specific to autoregressive
TTT: the difficulty of query estimation for shallow drafters, the
sparse gradient signal available to draft-side KV projections, and a
gate-induced gradient starvation effect in warm-started hybrid
drafters, where the learned gate closes early to suppress the
randomly-initialized branch and then struggles to reopen.  All three
stem from the sequential, token-by-token nature of autoregressive
drafting.  Block-wise training pipelines, which generate many draft
tokens in parallel, may be a better fit for these limitations and deserve direct evaluation in future KV-aware drafters.

The diagnostic value of this work lies in separating \emph{what}
information is available from \emph{whether} the training pipeline
can learn to use it.  KV cache does carry a useful signal for
long-range drafting; the challenge is building training setups that
can exploit it efficiently.

\newpage
\bibliographystyle{iclr2026_conference}
\bibliography{references/references}

@article{xia2024survey,
  author = {Xia, Heming and Yang, Zhe and Dong, Qingxiu and Wang, Peiyi and
            Li, Yongqi and Ge, Tao and Liu, Tianyu and Li, Wenjie and
            Sui, Zhifang},
  title = {Unlocking Efficiency in Large Language Model Inference: A
           Comprehensive Survey of Speculative Decoding},
  journal = {arXiv preprint arXiv:2401.07851},
  year = {2024},
  url = {https://arxiv.org/abs/2401.07851}
}

@article{deepseekv3,
  author = {{DeepSeek-AI}},
  title = {DeepSeek-V3 Technical Report},
  journal = {arXiv preprint arXiv:2412.19437},
  year = {2024},
  url = {https://arxiv.org/abs/2412.19437}
}

@article{chen2026dflash,
  author = {Chen, Jian and Liang, Yesheng and Liu, Zhijian},
  title = {DFlash: Block Diffusion for Flash Speculative Decoding},
  journal = {arXiv preprint arXiv:2602.06036},
  year = {2026},
  url = {https://arxiv.org/abs/2602.06036}
}

@misc{qwen-3,
  title={Qwen3 Technical Report},
  author={{Qwen Team} and An Yang and Anfeng Li and Baosong Yang and Beichen Zhang and Binyuan Hui and Bo Zheng and Bowen Yu and Chang Gao and Chengen Huang and Chenxu Lv and Chujie Zheng and Dayiheng Liu and Fan Zhou and Fei Huang and Feng Hu and Hao Ge and Haoran Wei and Huan Lin and Jialong Tang and Jian Yang and Jianhong Tu and Jianwei Zhang and Jianxin Yang and Jiaxi Yang and Jing Zhou and Jingren Zhou and Junyang Lin and Kai Dang and Keqin Bao and Kexin Yang and Le Yu and Lianghao Deng and Mei Li and Mingfeng Xue and Mingze Li and Pei Zhang and Peng Wang and Qin Zhu and Rui Men and Ruize Gao and Shixuan Liu and Shuang Luo and Tianhao Li and Tianyi Tang and Wenbiao Yin and Xingzhang Ren and Xinyu Wang and Xinyu Zhang and Xuancheng Ren and Yang Fan and Yang Su and Yichang Zhang and Yinger Zhang and Yu Wan and Yuqiong Liu and Zekun Wang and Zeyu Cui and Zhenru Zhang and Zhipeng Zhou and Zihan Qiu},
  year={2025},
  eprint={2505.09388},
  archivePrefix={arXiv},
  primaryClass={cs.CL},
  url={https://arxiv.org/abs/2505.09388}
}

@misc{mtp,
  title={Better \& Faster Large Language Models via Multi-token Prediction},
  author={Fabian Gloeckle and Badr Youbi Idrissi and Baptiste Rozi{\`e}re and David Lopez-Paz and Gabriel Synnaeve},
  year={2024},
  eprint={2404.19737},
  archivePrefix={arXiv},
  primaryClass={cs.CL},
  url={https://arxiv.org/abs/2404.19737}
}

@InProceedings{sps1,
  title = {Fast Inference from Transformers via Speculative Decoding},
  author = {Leviathan, Yaniv and Kalman, Matan and Matias, Yossi},
  booktitle = {Proceedings of the 40th International Conference on Machine Learning},
  pages = {19274--19286},
  year = {2023},
  volume = {202},
  series = {Proceedings of Machine Learning Research},
  month = {23--29 Jul},
  publisher = {PMLR},
  url = {https://proceedings.mlr.press/v202/leviathan23a.html}
}

@misc{sps2,
  title={Accelerating Large Language Model Decoding with Speculative Sampling},
  author={Charlie Chen and Sebastian Borgeaud and Geoffrey Irving and Jean-Baptiste Lespiau and Laurent Sifre and John Jumper},
  year={2023},
  eprint={2302.01318},
  archivePrefix={arXiv},
  primaryClass={cs.CL},
  url={https://arxiv.org/abs/2302.01318}
}

@misc{pearl,
  title={PEARL: Parallel Speculative Decoding with Adaptive Draft Length},
  author={Tianyu Liu and Yun Li and Qitan Lv and Kai Liu and Jianchen Zhu and Winston Hu and Xiao Sun},
  year={2025},
  eprint={2408.11850},
  archivePrefix={arXiv},
  primaryClass={cs.CL},
  url={https://arxiv.org/abs/2408.11850}
}

@inproceedings{specinfer,
  author = {Miao, Xupeng and Oliaro, Gabriele and Zhang, Zhihao and Cheng, Xinhao and Wang, Zeyu and Zhang, Zhengxin and Wong, Rae Ying Yee and Zhu, Alan and Yang, Lijie and Shi, Xiaoxiang and Shi, Chunan and Chen, Zhuoming and Arfeen, Daiyaan and Abhyankar, Reyna and Jia, Zhihao},
  title = {SpecInfer: Accelerating Large Language Model Serving with Tree-based Speculative Inference and Verification},
  year = {2024},
  isbn = {9798400703867},
  publisher = {Association for Computing Machinery},
  address = {New York, NY, USA},
  url = {https://doi.org/10.1145/3620666.3651335},
  doi = {10.1145/3620666.3651335},
  booktitle = {Proceedings of the 29th ACM International Conference on Architectural Support for Programming Languages and Operating Systems, Volume 3},
  pages = {932--949},
  numpages = {18},
  location = {La Jolla, CA, USA},
  series = {ASPLOS '24}
}

@inproceedings{medusa,
  author = {Cai, Tianle and Li, Yuhong and Geng, Zhengyang and Peng, Hongwu and Lee, Jason D. and Chen, Deming and Dao, Tri},
  title = {MEDUSA: Simple LLM inference acceleration framework with multiple decoding heads},
  year = {2024},
  publisher = {JMLR.org},
  booktitle = {Proceedings of the 41st International Conference on Machine Learning},
  articleno = {203},
  numpages = {27},
  location = {Vienna, Austria},
  series = {ICML'24},
  url = {https://proceedings.mlr.press/v235/cai24b.html}
}

@misc{eagle,
  title={EAGLE: Speculative Sampling Requires Rethinking Feature Uncertainty},
  author={Yuhui Li and Fangyun Wei and Chao Zhang and Hongyang Zhang},
  year={2025},
  eprint={2401.15077},
  archivePrefix={arXiv},
  primaryClass={cs.LG},
  url={https://arxiv.org/abs/2401.15077}
}

@inproceedings{eagle2,
  title = "{EAGLE}-2: Faster Inference of Language Models with Dynamic Draft Trees",
  author = "Li, Yuhui and Wei, Fangyun and Zhang, Chao and Zhang, Hongyang",
  booktitle = "Proceedings of the 2024 Conference on Empirical Methods in Natural Language Processing",
  month = nov,
  year = "2024",
  address = "Miami, Florida, USA",
  publisher = "Association for Computational Linguistics",
  url = "https://aclanthology.org/2024.emnlp-main.422/",
  doi = "10.18653/v1/2024.emnlp-main.422",
  pages = "7421--7432"
}

@misc{eagle3,
  title={EAGLE-3: Scaling up Inference Acceleration of Large Language Models via Training-Time Test},
  author={Yuhui Li and Fangyun Wei and Chao Zhang and Hongyang Zhang},
  year={2025},
  eprint={2503.01840},
  archivePrefix={arXiv},
  primaryClass={cs.CL},
  url={https://arxiv.org/abs/2503.01840}
}

@inproceedings{lade,
  author = {Fu, Yichao and Bailis, Peter and Stoica, Ion and Zhang, Hao},
  title = {Break the sequential dependency of LLM inference using LOOKAHEAD DECODING},
  year = {2024},
  publisher = {JMLR.org},
  booktitle = {Proceedings of the 41st International Conference on Machine Learning},
  articleno = {561},
  numpages = {20},
  location = {Vienna, Austria},
  series = {ICML'24},
  url = {https://proceedings.mlr.press/v235/fu24a.html}
}

@inproceedings{glide,
  author = {Du, Cunxiao and Jiang, Jing and Xu, Yuanchen and Wu, Jiawei and Yu, Sicheng and Li, Yongqi and Li, Shenggui and Xu, Kai and Nie, Liqiang and Tu, Zhaopeng and You, Yang},
  title = {GLIDE with a CAPE: a low-hassle method to accelerate speculative decoding},
  year = {2024},
  publisher = {JMLR.org},
  booktitle = {Proceedings of the 41st International Conference on Machine Learning},
  articleno = {465},
  numpages = {17},
  location = {Vienna, Austria},
  series = {ICML'24}
}

@misc{longspec,
  title={LongSpec: Long-Context Lossless Speculative Decoding with Efficient Drafting and Verification},
  author={Penghui Yang and Cunxiao Du and Fengzhuo Zhang and Haonan Wang and Tianyu Pang and Chao Du and Bo An},
  year={2026},
  eprint={2502.17421},
  archivePrefix={arXiv},
  primaryClass={cs.CL},
  note={Accepted to ACL 2026},
  url={https://arxiv.org/abs/2502.17421}
}

@inproceedings{Zhang_2024,
  title = {Draft\& Verify: Lossless Large Language Model Acceleration via Self-Speculative Decoding},
  url = {http://dx.doi.org/10.18653/v1/2024.acl-long.607},
  DOI = {10.18653/v1/2024.acl-long.607},
  booktitle = {Proceedings of the 62nd Annual Meeting of the Association for Computational Linguistics (Volume 1: Long Papers)},
  publisher = {Association for Computational Linguistics},
  author = {Zhang, Jun and Wang, Jue and Li, Huan and Shou, Lidan and Chen, Ke and Chen, Gang and Mehrotra, Sharad},
  year = {2024},
  pages = {11263--11282}
}

@misc{lv2026hippoacceleratingvideolarge,
  title={HIPPO: Accelerating Video Large Language Models Inference via Holistic-aware Parallel Speculative Decoding},
  author={Qitan Lv and Tianyu Liu and Wen Wu and Xuenan Xu and Bowen Zhou and Feng Wu and Chao Zhang},
  year={2026},
  eprint={2601.08273},
  archivePrefix={arXiv},
  primaryClass={cs.CV},
  url={https://arxiv.org/abs/2601.08273}
}

@misc{xia2025swiftontheflyselfspeculativedecoding,
  title={SWIFT: On-the-Fly Self-Speculative Decoding for LLM Inference Acceleration},
  author={Heming Xia and Yongqi Li and Jun Zhang and Cunxiao Du and Wenjie Li},
  year={2025},
  eprint={2410.06916},
  archivePrefix={arXiv},
  primaryClass={cs.CL},
  url={https://arxiv.org/abs/2410.06916}
}

@inproceedings{song-etal-2026-knn,
  title = "{KNN}-{SSD}: Enabling Dynamic Self-Speculative Decoding via Nearest Neighbor Layer Set Optimization",
  author = "Song, Mingbo and
    Xia, Heming and
    Zhang, Jun and
    Leong, Chak Tou and
    Xu, Qiancheng and
    Li, Wenjie and
    Li, Sujian",
  editor = "Demberg, Vera and
    Inui, Kentaro and
    Marquez, Llu{\'i}s",
  booktitle = "Findings of the {Association for Computational Linguistics}: {EACL} 2026",
  month = mar,
  year = "2026",
  address = "Rabat, Morocco",
  publisher = "Association for Computational Linguistics",
  url = "https://aclanthology.org/2026.findings-eacl.31/",
  doi = "10.18653/v1/2026.findings-eacl.31",
  pages = "641--655",
  ISBN = "979-8-89176-386-9"
}

@misc{sequoia,
  title={Sequoia: Scalable, Robust, and Hardware-aware Speculative Decoding},
  author={Zhuoming Chen and Avner May and Ruslan Svirschevski and Yuhsun Huang and Max Ryabinin and Zhihao Jia and Beidi Chen},
  year={2025},
  eprint={2402.12374},
  archivePrefix={arXiv},
  primaryClass={cs.CL},
  url={https://arxiv.org/abs/2402.12374}
}

@inproceedings{hass,
  title={Learning Harmonized Representations for Speculative Sampling},
  author={Lefan Zhang and Xiaodan Wang and Yanhua Huang and Ruiwen Xu},
  booktitle={The Thirteenth International Conference on Learning Representations},
  year={2025},
  url={https://openreview.net/forum?id=T9u56s7mbk}
}

@misc{logitspec,
  title={LogitSpec: Accelerating Retrieval-based Speculative Decoding via Next Next Token Speculation},
  author={Tianyu Liu and Qitan Lv and Hao Li and Xing Gao and Xiao Sun and Xiaoyan Sun},
  year={2026},
  eprint={2507.01449},
  archivePrefix={arXiv},
  primaryClass={cs.CL},
  url={https://arxiv.org/abs/2507.01449}
}

@misc{talon,
  title={TALON: Confidence-Aware Speculative Decoding with Adaptive Token Trees},
  author={Tianyu Liu and Qitan Lv and Yuhao Shen and Xiao Sun and Xiaoyan Sun},
  year={2026},
  eprint={2601.07353},
  archivePrefix={arXiv},
  primaryClass={cs.CL},
  url={https://arxiv.org/abs/2601.07353}
}

@misc{specbranch,
  title={SpecBranch: Speculative Decoding via Hybrid Drafting and Rollback-Aware Branch Parallelism},
  author={Yuhao Shen and Junyi Shen and Quan Kong and Tianyu Liu and Yao Lu and Cong Wang},
  year={2026},
  eprint={2506.01979},
  archivePrefix={arXiv},
  primaryClass={cs.DC},
  url={https://arxiv.org/abs/2506.01979}
}

@misc{double,
  title={Double: Breaking the Acceleration Limit via Double Retrieval Speculative Parallelism},
  author={Yuhao Shen and Tianyu Liu and Junyi Shen and Jinyang Wu and Quan Kong and Li Huan and Cong Wang},
  year={2026},
  eprint={2601.05524},
  archivePrefix={arXiv},
  primaryClass={cs.CL},
  url={https://arxiv.org/abs/2601.05524}
}

@misc{specforge2025,
  title={SpecForge: Train speculative decoding models effortlessly},
  author={Shenggui Li and Yikai Zhu and Chao Wang and Fan Yin and Shuai Shi and Yubo Wang and Yi Zhang and Yingyi Huang and Haoshuai Zheng and Yineng Zhang},
  year={2025},
  publisher={GitHub},
  howpublished={\url{https://github.com/sgl-project/specforge}}
}

@misc{sharegpt-vicuna-unfiltered,
  title = {{ShareGPT}},
  author = {{ShareGPT}},
  year = {2023},
  howpublished = {\url{https://huggingface.co/datasets/Aeala/ShareGPT_Vicuna_unfiltered}}
}

\newpage
\appendix
\section{Training Details}
\label{app:training}

All drafter models are trained using the autoregressive test-time
training (TTT) objective of EAGLE-3 \citep{eagle3}.  The target model
is Qwen3-8B \citep{qwen-3} throughout.

\paragraph{Target KV layers.}
Unless otherwise noted, the KVShot testbed samples KV cache from $L_s = 3$
uniformly spaced target layers, matching the number of layers fused
by EAGLE-3.  Section~\ref{app:ablations} shows that varying this
choice has negligible effect.

\paragraph{Data.}
Ablation experiments (Sections~\ref{sec:pure-kv}--\ref{sec:hybrid})
train on ShareGPT (${\sim}$70k samples, 3~epochs).  The end-to-end
evaluation (Section~\ref{sec:e2e}) scales to 280k samples
(ShareGPT~+~UltraChat) with responses regenerated by the target model
to ensure distributional alignment.

\paragraph{Evaluation.}
We compute step-wise acceptance rates $\alpha_k$ on a held-out subset.
We measure end-to-end MAT using the HuggingFace speculative decoding
pipeline with a draft tree configuration of $(8, 10, 60)$.

% ------------------------------------------------------------------
\section{Gated KV Fusion: Architecture Details}
\label{app:architecture}

This section expands on the gated delta fusion design introduced in
Section~\ref{sec:hybrid}.

\paragraph{Input representation.}
Each draft token's input $x_i \in \mathbb{R}^d$ follows the EAGLE-3
construction: the input embedding and target hidden states are
concatenated and projected.  We do not modify this input pathway; all
changes are confined to the attention sub-module within the single
draft layer.

\paragraph{Cross-attention memory.}
The cross-attention branch attends to a mixed memory
$\mathcal{M}_i$ that changes per draft step.  For verified prefix
tokens ($j \leq t$), $\mathcal{M}_i$ contains target KV pairs copied
directly from the target model.  For subsequent draft positions
($j > t$), the drafter generates its own KV via learned projections
$W_K^{\mathrm{cross}}$ and $W_V^{\mathrm{cross}}$.  This design
mirrors the standard autoregressive KV cache structure: the prefix
portion is ``free'' (no estimation needed), while draft-generated
keys and values must be learned.

\paragraph{Why a gated delta, not a direct sum.}
A naive residual combination $h_i = n_i + o_i$ allows the
cross-attention branch to override the self-attention output,
potentially destroying the short-range accuracy that the EAGLE-3 path
provides.  The gated delta formulation
(\autoref{eq:delta}--\autoref{eq:fusion}) instead treats the
cross-attention output as a \emph{correction}: the difference
$\Delta_i = o_i - n_i$ is first projected by $W_{\Delta}$, then
scaled element-wise by a learned gate $g_i \in [0,1]^d$.  When
$g_i \approx 0$, the output collapses to the self-attention path;
the cross-attention branch contributes only where the gate deems the
correction beneficial.

\paragraph{Shared gate.}
The gate parameters $W_g \in \mathbb{R}^{d \times 3d}$ and
$b_g \in \mathbb{R}^{d}$ are shared across all draft steps.
We chose a shared gate over per-step gates for two reasons:
(1)~the primary question at this stage is whether the gated delta
mechanism is effective at all, not whether step-specific gating
improves it further; and (2)~sharing reduces parameter count and
stabilizes training, which matters when the cross-attention branch is
randomly initialized.

\paragraph{Initialization from EAGLE-3 checkpoint.}
When warm-starting from a pre-trained EAGLE-3 checkpoint, the
self-attention branch and all shared modules (input projection, MLP,
layer norms) load existing weights directly.  The cross-attention
projections ($W_Q^{\mathrm{cross}}$, $W_K^{\mathrm{cross}}$,
$W_V^{\mathrm{cross}}$, $W_O^{\mathrm{cross}}$), the gate ($W_g$,
$b_g$), and the delta projection ($W_{\Delta}$) are randomly
initialized.  This allows the model to begin training from a strong
self-attention anchor while gradually learning the cross-attention
correction.

For the \emph{Cross-only} variant in Table~\ref{tab:hybrid}, the
EAGLE-3-checkpoint setting uses the same warm-start scheme: the
inherited query/key/value/output projections and MLP are loaded from
the EAGLE-3 checkpoint, but the self-attention output $n_i$ is
omitted from the forward pass so that $h_i$ depends only on the
cross-attention branch.  This isolates the contribution of the
hidden-state pathway from the choice of initialization.

% ------------------------------------------------------------------
\section{Full Step-wise Acceptance Rates}
\label{app:full-rates}

Tables~\ref{tab:pure-kv}--\ref{tab:hybrid} in the main text report
representative checkpoints rather than the full 7-step curves.
Table~\ref{tab:full-rates} provides the complete 7-step acceptance
rates ($\alpha_0$ through $\alpha_6$) for all experiments in
the main tables.

\begin{table}[h]
\centering
\caption{Complete step-wise acceptance rates for experiments reported
    in the main text.  Group headings correspond to the main tables.}
\label{tab:full-rates}
\small
\setlength{\tabcolsep}{4pt}
\begin{tabular}{@{}lcccccccr@{}}
\toprule
Method
    & $\alpha_0$ & $\alpha_1$ & $\alpha_2$ & $\alpha_3$
    & $\alpha_4$ & $\alpha_5$ & $\alpha_6$ & MAT \\
\midrule
\multicolumn{9}{l}{\textit{Pure KV reuse (1-layer, Table~\ref{tab:pure-kv})}} \\
No target info
    & .237 & .238 & .237 & .237 & .237 & .237 & .236 & 1.31 \\
Head concat
    & .489 & .393 & .353 & .332 & .320 & .311 & .305 & 1.78 \\
Linear projection
    & .488 & .425 & .396 & .379 & .368 & .360 & .354 & 1.83 \\
Proj + RoPE fix
    & .494 & .427 & .397 & .381 & .369 & .360 & .353 & 1.84 \\
\midrule
\multicolumn{9}{l}{\textit{Depth scaling (Table~\ref{tab:depth})}} \\
Pure KV 1-layer
    & .494 & .427 & .397 & .381 & .369 & .360 & .353 & 1.84 \\
Pure KV 2-layer
    & .594 & .540 & .514 & .497 & .484 & .473 & .463 & 2.23 \\
Pure KV 3-layer
    & .609 & .558 & .534 & .518 & .506 & .496 & .487 & 2.31 \\
Pure KV 4-layer
    & .614 & .565 & .542 & .527 & .515 & .505 & .495 & 2.34 \\
\midrule
\multicolumn{9}{l}{\textit{Gated KV fusion (1-layer, Table~\ref{tab:hybrid})}} \\
Gated KV (scratch)
    & .650 & .582 & .548 & .527 & .512 & .501 & .490 & 2.44 \\
Gated KV (ckpt)
    & .665 & .603 & .573 & .553 & .537 & .525 & .514 & 2.54 \\
Cross-only (scratch)
    & .634 & .561 & .523 & .499 & .480 & .464 & .450 & 2.34 \\
Cross-only (ckpt)
    & .637 & .563 & .522 & .495 & .475 & .458 & .442 & 2.35 \\
\midrule
\multicolumn{9}{l}{\textit{Baseline}} \\
EAGLE-3 (1-layer)
    & .638 & .566 & .533 & .511 & .495 & .481 & .469 & 2.37 \\
\bottomrule
\end{tabular}
\end{table}

Two patterns become clearer with the full curves.
First, the decay profile of the \textbf{Gated KV (ckpt)} model is
uniformly above the EAGLE-3 baseline at every step, showing that the
improvement is not concentrated at a single step.
Second, the \textbf{Head concat} variant decays markedly faster
than \textbf{Linear projection} at early steps
($\alpha_1$: 0.393 vs.\ 0.425), indicating that the difficulty of
attending across $L_s$ disjoint KV spaces manifests immediately
rather than only at long range.

% ------------------------------------------------------------------
\section{Additional Ablations}
\label{app:ablations}

Table~\ref{tab:ablations} collects ablations referenced in the main
text but omitted from the main tables.  All rows use the 2-layer pure
KV drafter with linear projection and RoPE re-application as the
baseline (Section~\ref{sec:depth-scaling}).

\begin{table}[h]
\centering
\caption{Additional ablations on 2-layer pure KV reuse (ShareGPT
    training).  The first row repeats the 2-layer baseline from
    Table~\ref{tab:depth} for reference.  `Retention' denotes the
    long-range retention ratio $\alpha_6 / \alpha_0$.}
\label{tab:ablations}
\small
\begin{tabular}{@{}lccccc@{}}
\toprule
Variant & $\alpha_0$ & $\alpha_3$ & $\alpha_6$ & Retention & MAT \\
\midrule
\multicolumn{6}{l}{\textit{Baseline}} \\
2-layer pure KV (online, linear proj)
    & 0.594 & 0.497 & 0.463 & 77.9\% & 2.23 \\
\midrule
\multicolumn{6}{l}{\textit{Target KV configuration}} \\
4 KV layers (vs.\ 3)
    & 0.600 & 0.495 & 0.460 & 76.7\% & 2.24 \\
\midrule
\multicolumn{6}{l}{\textit{Projector architecture}} \\
MLP projector
    & 0.586 & 0.491 & 0.458 & 78.2\% & 2.20 \\
MLP projector + LayerNorm
    & 0.583 & 0.491 & 0.458 & 78.6\% & 2.19 \\
Hidden$\to$KV cross-projection
    & 0.558 & 0.429 & 0.388 & 69.5\% & 2.04 \\
\midrule
\multicolumn{6}{l}{\textit{Training regime}} \\
QK-Norm
    & 0.596 & 0.498 & 0.466 & 78.2\% & 2.23 \\
Offline TTT
    & 0.582 & 0.488 & 0.454 & 78.0\% & 2.18 \\
Offline TTT + KV grad scale ($50\times$)
    & 0.587 & 0.493 & 0.459 & 78.2\% & 2.21 \\
\bottomrule
\end{tabular}
\end{table}

\paragraph{Target KV configuration.}
Increasing the number of sampled target layers from 3 to 4 yields
only a marginal gain (MAT~2.24 vs.\ 2.23), indicating that the
drafter's bottleneck lies in query estimation rather than in the
richness of the KV input.

\paragraph{Projector architecture.}
Replacing the linear projector with a two-layer MLP (with or without
LayerNorm) slightly hurts performance (MAT~2.20 and 2.19 vs.\ 2.23).
The added expressiveness does not offset the optimization difficulty,
consistent with the sparse-gradient analysis in
Section~\ref{sec:kv-gradient}.  A more revealing variant---projecting
target hidden states into KV space rather than directly reusing
pre-computed KV pairs---yields a substantially lower MAT~(2.04),
suggesting that the pre-computed KV representation itself matters.

\paragraph{Training regime.}
Switching from online TTT to offline training---i.e., conditioning on
target hidden states without autoregressive drift alignment---reduces
MAT from 2.23 to 2.18, indicating that TTT's drift alignment benefits
even KV-based drafters.  Scaling the KV-projection gradient by
$50\times$ recovers only a small portion of the gap (MAT~2.21),
reinforcing the conclusion from Section~\ref{sec:kv-gradient} that the
core issue is training signal diversity, not gradient magnitude.
QK-Norm produces no measurable change (MAT~2.23).

\end{document}